\documentclass{bmcart}
\usepackage{graphicx}
\usepackage{setspace}
\usepackage{amsmath}
\usepackage{booktabs}
\usepackage{adjustbox}
\usepackage[numbers,sort&compress]{natbib}

\usepackage{amsthm,amsmath}
\usepackage[utf8]{inputenc} 

\startlocaldefs
\endlocaldefs

\begin{document}

\begin{frontmatter}

\begin{fmbox}
\dochead{Research}


\title{Predicting multiple sclerosis disease severity with multimodal deep neural networks}


\author[
  addressref={aff1},                   
  email={kai.zhang.1@uth.tmc.edu}      
]{\inits{K.Z.}\fnm{Kai} \snm{Zhang}}
\author[
  addressref={aff2},
  email={john.a.lincoln@uth.tmc.edu}
]{\inits{J.L.} \fnm{John A.} \snm{Lincoln}}
\author[
  addressref={aff1},
  email={xiaoqian.jiang@uth.tmc.edu}
]{\inits{X.J.}\fnm{Xiaoqian} \snm{Jiang}}
\author[
  addressref={aff1, aff3},
  email={Elmer.V.Bernstam@uth.tmc.edu}
]{\inits{E.B.}\fnm{Elmer V.} \snm{Bernstam}}
\author[
  addressref={aff1, aff4},
  corref={aff4},
  email={Shayan.Shams@sjsu.edu}
]{\inits{S.S.}\fnm{Shayan} \snm{Shams}}


\address[id=aff1]{
  \orgdiv{School of Biomedical Informatics},             
  \orgname{University of Texas Health Sciences Center at Houston},          
  \city{Houston, TX},                              
  \cny{United States}                                    
}
\address[id=aff2]{%
  \orgdiv{Department of Neurology},
  \orgname{University of Texas Health Sciences Center, McGovern Medical School},
  \city{Houston, TX},                              
  \cny{United States}  
}

\address[id=aff3]{%
  \orgdiv{Division of General Internal Medicine, Department of Internal Medicine},
  \orgname{University of Texas Health Sciences Center, McGovern Medical School},
  \city{Houston, TX},                              
  \cny{United States}  
}

\address[id=aff4]{%
  \orgdiv{Department of Applied Data Science},
  \orgname{San Jose State University},
  \city{San Jose, CA},                              
  \cny{United States}  
}


\end{fmbox}


\begin{abstractbox}

\begin{abstract} 
Multiple Sclerosis (MS) is a chronic disease developed in human brain and spinal cord, which can cause permanent damage or deterioration of the nerves. The severity of MS disease is monitored by the Expanded Disability Status Scale (EDSS), composed of several functional sub-scores. Early and accurate classification of MS disease severity is critical for slowing down or preventing disease progression via applying early therapeutic intervention strategies. Recent advances in deep learning and the wide use of Electronic Health Records (EHR) creates opportunities to apply data-driven and predictive modeling tools for this goal. Previous studies focusing on using single-modal machine learning and deep learning algorithms were limited in terms of prediction accuracy due to the data insufficiency or model simplicity. In this paper, we proposed an idea of using patients' multimodal longitudinal and longitudinal EHR data to predict multiple sclerosis disease severity at the hospital visit. This work has two important contributions. First, we describe a pilot effort to leverage structured EHR data, neuroimaging data and clinical notes to build a multi-modal deep learning framework to predict patient's MS disease severity. The proposed pipeline demonstrates up to 25\% increase in terms of the area under the Area Under the Receiver Operating Characteristic curve (AUROC) compared to models using single-modal data. Second, the study also provides insights regarding the amount useful signal embedded in each data modality with respect to MS disease prediction, which may improve data collection processes.
\end{abstract}




\end{abstractbox}
%

\end{frontmatter}




\section*{Background}
Recent advantages in deep learning have shown success in various areas of healthcare, such as brain Magnetic Resonance Imaging (MRI) automatic volume segmentation and classification \cite{korolev2017residual}, clinical text mining and disease prediction \cite{costa2020multiple}, risk predictions \cite{zhang2021real}, etc. The fast-growing Electronic Health Records (EHR) in healthcare provides a great number of opportunities for both the data mining and deep learning communities to explore the rich information embedded in different data modalities and tap the potentiality of using this information for predictive modeling, to benefit effective healthcare delivery and better-quality caring for patients. 

Multiple sclerosis (MS) is a potentially disabling disease that affects the human brain and spinal cord. An estimation of MS prevalence by the year 2010 of 10-year accumulation shows there are over 700,000 MS cases in adults in the United States \cite{wallin2019prevalence}. Recent advantages in MS disease research found that patients who died from MS suffer up to 39\% neuron count loss compared to usual patients without MS \cite{carassiti2018neuronal}. The human brain has mechanisms for self-repair and regenerative potential that could repair the brain plaques \cite{charles2002re}, however, such ability is very limited. Therefore, prompt action to prevent or slow down brain damage is critical to MS disease treatment \cite{giovannoni2016brain}. Effective treatment relies on a correct grading of the MS severity, and scoring systems are widely used to achieve this goal. The Expanded Disability Status Scale (EDSS) score \cite{kurtzke1983rating} is a widely used ordinal scoring system by healthcare providers to monitor clinical disability in MS. It is composed of diverse functional systems, including pyramidal functions (muscle strength, tone, and reflexes), cerebellar functions (coordination and balance), brainstem functions (eye movements, speech, and swallowing), sensory functions (light touch, pain, and vibratory sense), bowel and bladder functions, visual functions, cerebral (cognition) and ambulation. Based on EDSS, Roxburgh et al. proposed a Multiple Sclerosis Severity Score (MSSS) which can be used to determine MS disease progression using single assessment data (when a patient has only one assessment during the disease course) \cite{roxburgh2005multiple}.

Several milestones of the EDSS score have been commonly used to define different stages of the MS disease course. The EDSS 4 (significant disability but able to walk without aid or rest for 500 m), EDSS 6 (requires unilateral assistance to walk about 100 m with or without resting) and EDSS 7 (ability to walk no more than 10 m without rest while leaning against a wall or holding onto furniture for support) were commonly-used milestones for studying MS disease severity. For example, Confavreux et al. used the above milestones to study the effect of relapses on the progression of irreversible disability \cite{confavreux2000relapses}. The same milestones have also been used to study the contribution of relapses to worsening disability and evaluate the MS therapies' effect on delaying the disability accumulation \cite{lublin2022patients}. A Sweden research group studied whether the risk of reaching the above disability milestones in MS has changed over the last decade \cite{beiki2019changes}. Rzepiński et al. used the EDSS milestones to explore early clinical features of MS and how they affect patients' long-term disability progression \cite{rzepinski2019early}. The same milestones were also used to study how these factors affect the time to transition from relapsing-remitting MS (RRMS) to secondary progressive MS (SPMS).

A patient's EDSS score needs to be evaluated by a well-trained specialist to ensure that the assessment is correctly performed, which limits the application of EDSS to clinics with MS disease specialties. Several research studies have attempted to address this problem using machine learning or deep learning models. In particular, Pinto et al. proposed to use machine learning models to predict MS progression, based on the clinical characteristics of the first five years of the disease \cite{pinto2020prediction}. Zhao et al. used a support vector machine (SVM) classifier and demographic, clinical, and MRI data obtained at years one and two to predict patients' EDSS at five years follow-ups \cite{zhao2017exploration}. Sacca et al. explored different machine learning models (Random Forest, Support Vector Machine, Naive-Bayes, K-nearest-neighbor, and Artificial Neural Network) and used the features extracted from functional-MRI to perform MS disease severity classification \cite{sacca2019evaluation}. Narayana et al. proposed to use the VGG-16 convolutional neural network (CNN) to predict enhancing lesions in MS patients using non-contrast MRIs \cite{narayana2020deep}. D'Costa et al. proposed a transformer model named MS-BERT to predict EDSS score from patient's neurological consult note \cite{costa2020multiple}. Ciotti proposed a clinical instrument to retrospectively capture levels of EDSS and the algorithm got a Kappa score of 0.80 between captured EDSS and real EDSS \cite{ciotti2020clinical}. Chase et al. also used neurological consult notes but with simpler models (Naïve Bayes classification model) and features (word frequency) \cite{chase2017early}. Dekker et al. used multiple linear regression models on patient brain lesion volumes and its variation over the years to predict physical disability \cite{dekker2019predicting}. The aforementioned studies explored the idea of using machine learning and deep learning methods on various modalities of EHR datasets to predict patient's EDSS of the current hospital visit or in the near future. The above works only explored a limited amount of patient information (either clinical notes, or basic lesion volume information extracted from MRI, or patient clinical characteristics), by adopting the off-the-shelf machine learning models, or deep learning models that were developed for general tasks and utilized without being customized to capture the complex nature of this prediction problem. Based on the above studies and the recent research advances in multimodal deep learning, it is reasonable to assume that using multimodal deep learning methods could integrate fragmented information from each modality and brings more accurate predictions for MS disease. Therefore, this study tried to answer the question of whether can we harmonize all the available EHR data modalities collected from patient clinic visits and use longitudinal data to perform more accurate MS severity prediction. A few research study findings have found that MRI data and some lab tests can contain useful information about MS disease severity. For example, studies have shown that the thickness of cortical and deep grey matter has a high correlation with the MS disease severity, suggesting that the MRI images are an informative data source to predict MS severity \cite{popescu2015drives, klaver2013grey}. Some laboratory tests were also documented as playing an important role in this regard, such as the cerebrospinal fluid (CSF) \cite{freedman2005recommended, thompson2018diagnosis}, and serum neurofilament light chain (nFl) \cite{disanto2017serum}.

This study tried to answer the above question using a data-driven approach. We explored the idea of using patients' MRI images, clinical notes, and structured EHR data (including laboratory tests, vital sign observations, medication prescriptions, patient demographics) that were collected during patients' clinic visits to predict MS disease severity at the visit. We propose a multimodal deep neural network that takes MS patient Electronic Health Records of multiple modalities, including the MRI images (pre- and post-contrast T1 weighted image, T2 weighted image, fluid-attenuated inversion recovery image and proton density image), patient's clinical notes data and structured EHR (laboratory tests, vital signs, medications, demographic information) to perform MS disease severity prediction. 

We also propose to use patients' longitudinal data for EDSS milestone prediction, based on the fact that evidence about patients' MS disease severity should not only be embedded in the most recent EHR data but also richly contained in data of all previous clinic visits. Compared to using cross-sectional data (e.g. using clinical notes of the current visit to predict EDSS score \cite{costa2020multiple}), we propose to use the patient's both current clinic visit and historical EHR data to train a multimodal deep neural network. The longitudinal data contains more MS disease progression information compared to cross-sectional data and will help the model make more accurate predictions of the patient's status at the moment. The contributions of this study are three-fold.

\begin{itemize}
    \item A novel deep learning architecture (a multimodal neural network) and data fusion mechanism which takes Electronic Health Records including medications, vital signs, laboratory test results, clinical imaging, and physician notes to tackle the difficult problem of MS disease severity prediction. The results show significant prediction accuracy improvements compared to using single-modality data or simpler models.
    \item Using longitudinal data (both current and historical visits data) instead of cross-sectional data (data of current visit) to accurately classify patient EDSS score milestones at the current clinic visit.
    \item Exhibits how much useful information is embedded in each data modality for the prediction of MS severity. Various attention mechanisms are adopted in the proposed neural network to provide model explainability and enhance prediction accuracy.
    \item An end-to-end AI model that works on readily available data with a limited pre-processing process (e.g. does not need feature extraction as a pre-processing step, such as extracting the thalamic volume, lateral ventricle volume, etc. to train the model) 

\end{itemize}

The paper is structured as follows. The Data section explains the dataset that was used in this study and the details of each data modality. The subsequent section explains our designed deep neural network architecture, followed by an Experiment section including our experiment design, the obtained results, and the discussion. Finally, a summary and conclusion are given in the Discussion section.

\section*{Data}
\begin{figure}[ht]
\centering
\includegraphics[width=0.95\textwidth]{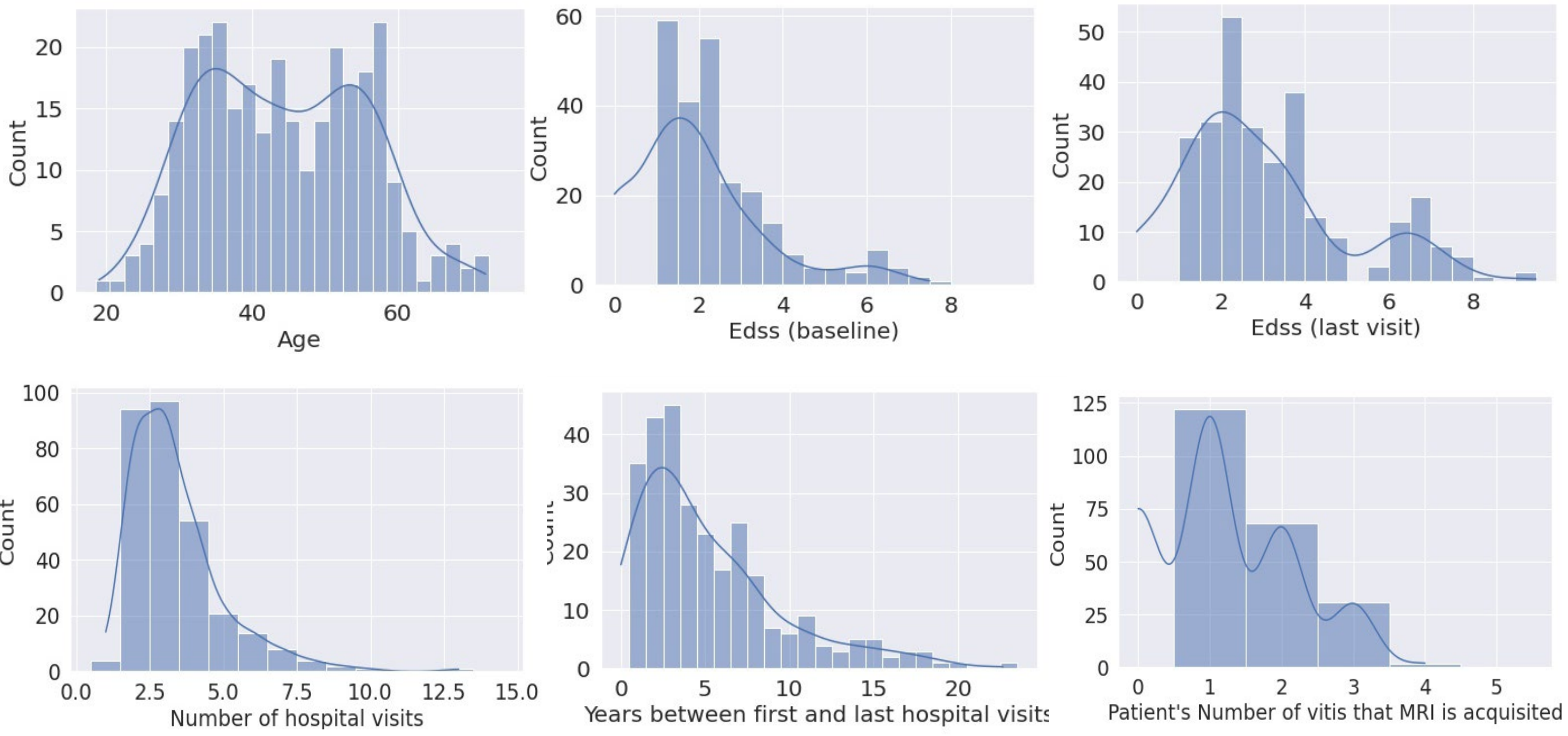}
\caption{The histograms of all patients by Age; Baseline EDSS (at initial hospital visit);  EDSS at the last hospital visit; Total hospital visits; Years between the first and the last hospital visit; Number of hospital visits during which brain MRI scan was performed.}
\label{figure:statistics}
\end{figure}

\begin{figure*}[tb]
    \centering
    \includegraphics[width=0.8\textwidth]{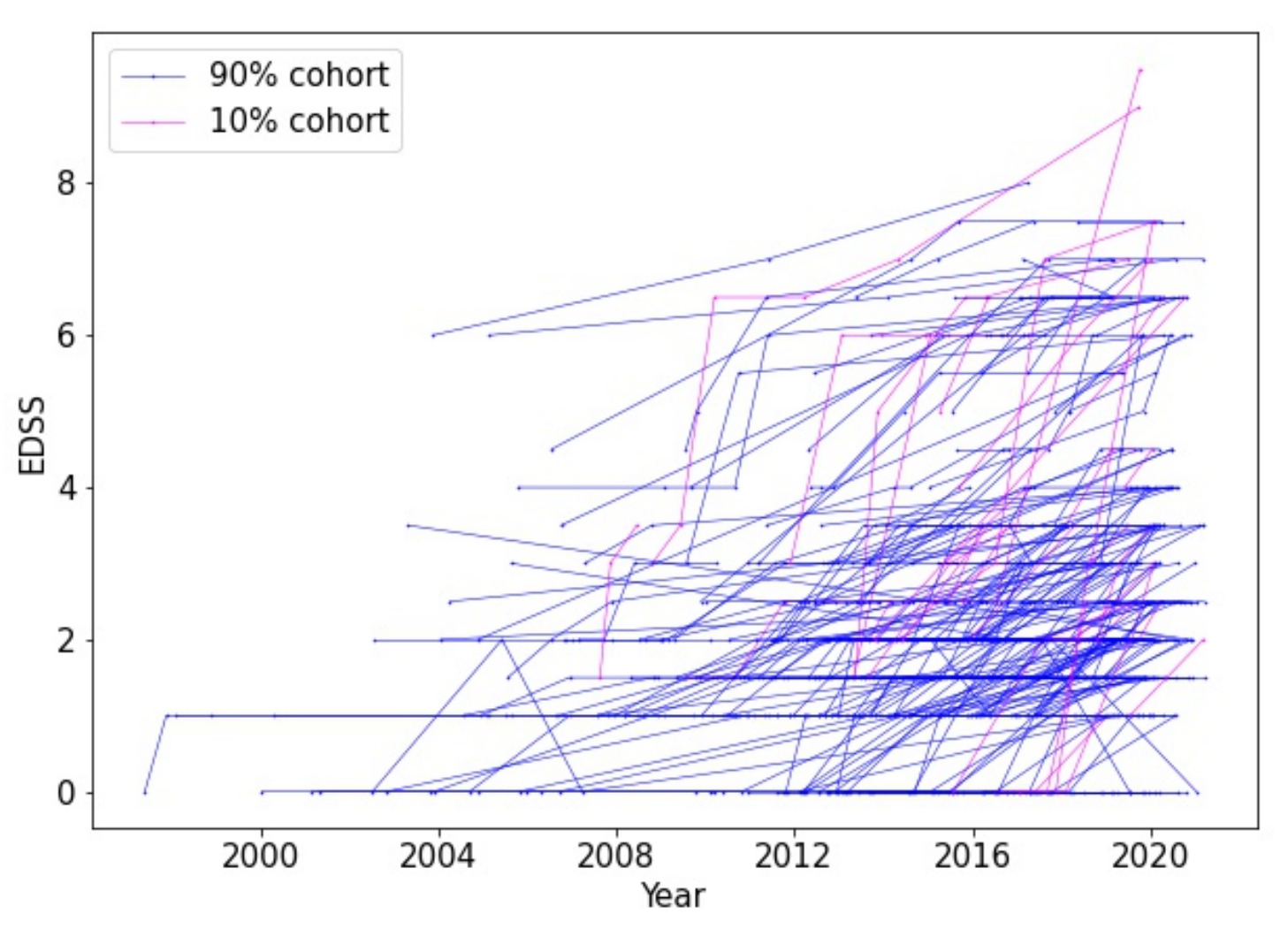}
    \caption{The MS disease progression of all patients. For clear illustration, patient were sorted by the total EDSS increase in their disease course and the trajectory of the top 10\% cohort who grows the most were marked in red, and the rest 90\% cohort were marked in blue.}
    \label{fig:progression}
\end{figure*}

\ref{sec-data}
Our database contains a rich set of 300 MS patients, patients' demographic information in summarized in Table \ref{table:statistics}. Each patient's data contain three modalities: 1) the neuroimaging data, 2) structured EHR data, and 3) clinical notes. The neuroimaging data is stored in NIFTI format. Most patients have multiple clinical visits, and during each clinic visit, a patient may have multiple laboratory tests, recorded vital signs, different prescription drugs, diagnoses, certain medical procedures, and treatments, that are recorded in the structured EHR data in separate tables. The clinical notes contain the physician's description of the patient's status at each clinic visit. Our proposed novel neural network architecture is designed to handle heterogeneous structure databases via learning representations of each modality. The prediction goal is set to a classification problem to predict if the patient has reached certain EDSS milestones at the current clinic visit. All 300 patient's EDSS scores were evaluated by physicians at the end of each clinic visit and were recorded in a table in the structured EHR. The real EDSS score was extracted from the EHR table and served as the ground truth label. The research goal is to develop a deep learning model to predict the EDSS score at the current visit using all other information and masking the true EDSS score. Figure \ref{figure:statistics} demonstrates the distributions of patients' age and EDSS. Figure \ref{fig:progression} plots all patients EDSS historical scores along their disease course.  

\begin{table} [ht]
\caption{An overview of patient statistics in the dataset (SD: standard deviation).}
\label{table:statistics}
\resizebox{\columnwidth}{!}{%
\begin{tabular}{|l|c|c|c|c|c|}
\hline
& Average $\pm$ SD & Minimum & Maximum & .25 quantile & .75 quantile \\ \hline
Age                        & 43.62 $\pm$ 11.20   & 19.00      & 71.00 & 34.00           & 52.00           \\ \hline
EDSS @ baseline             & 1.93 $\pm$  1.59   & 0   & 7.50    & 1.00  & 2.50          \\ \hline
EDSS @ last visit           & 2.90 $\pm$  1.96   & 0   & 9.50    & 1.50  & 3.50          \\ \hline
Number of visits       & 3.39 $\pm$ 1.60   & 1   & 13      & 2.00  & 4.00          \\ \hline
Years b/w first and last visits & 5.14  $\pm$  4.34  & 0   & 22.66   & 2.03  & 7.01         \\ \hline
Number of MRI sessions/patient  & 1.20  $\pm$  0.96  & 0   & 4       & 0     & 2            \\ \hline
\end{tabular}
}
\end{table}

\begin{figure}[ht]
\centering
\includegraphics[width=0.95\textwidth]{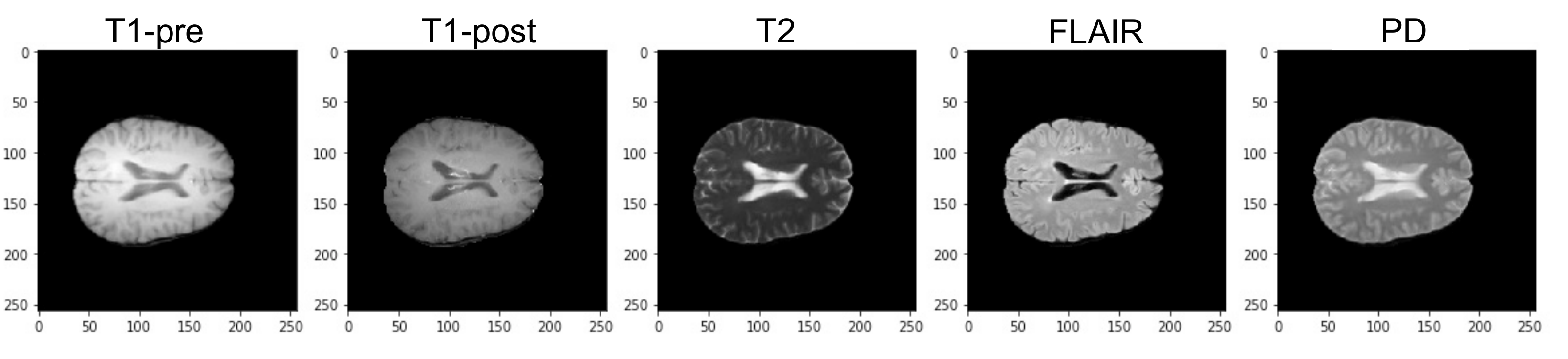}
\caption{The MRI sequences of a patient as an example.}
\label{fig:mri}
\end{figure}

\label{sec-data}
\textbf{Brain MRI.} We obtain a total of 360 MRI images for all 300 patients. All imaging studies were performed on a Philips 3.0T Ingenia scanner (Philips Medical Systems, Best, Netherlands). Some patients can have multiple MRIs from different clinic visits. The MRIs include five sequences: pre-contrast and post-contrast T1-weighted sequences (T1-pre, T1-post),  T2-weighted sequences, proton density-weighted sequences (PD) and fluid-attenuated inversion recovery sequences (FLAIR). All sequences were acquired with a field of view of 256 mm x 256 mm x 44 mm. For each patient, the MRI images were acquired in the axial plane. Figure \ref{fig:mri} displays the MRI sequences of a sample patient. All MRI sequences are skull-stripped using Simple Skull Stripping (S3) \cite{lipkova2019personalized} and the SRI24 template \cite{rohlfing2010sri24}, bias-corrected using N4 Bias Field Correction to adjust the low-frequency intensity \cite{tustison2010n4itk}, and  co-registered using FreeSurfer \cite{fischl2012freesurfer} to a common template (SRI24).

\textbf{Clinical Notes.} Patient's clinical notes are in free text format and contain the physician's description of the patient's health status, patient basic health information such as weight, height, BMI (body mass index), physiological status, diagnosis, medications and received treatments. We de-identified all clinical notes data by removing patients' and Physician's personal data.

\textbf{Structured EHR.}
Patient's structured EHR contains laboratory tests measurements (float), vital sign observations (float), medication administrations (0/1 indicator - taken or not taken), and demographic information (age: float, race/ethnicity/gender: 0/1) in a tabular format. We construct the tables in the format of rows being observational time stamps and columns representing a number of features. The features in each table are fixed for all patients, and the number of rows for different tables and different patients is different depending on how many observational time points a patient has. For each patient's laboratory test, vital signs,s and medication table, we set the time granularity to be 4 hours. We record each feature's average value if it has multiple observations during a 4-hour window. This helps to reduce table dimensions, eliminate data observational noises, and avoid creating large and sparse tables which impedes neural network training. If certain features have no values during the four-hour windows, the corresponding entry will be zero. The window size is treated as a hyper-parameter to be optimized to strike a balance between table dimension and information loss (a large averaging window smooths feature observations and blurs useful information, and a small window makes the table have a high dimension on the time-axis which and decrease the efficiency of network training). The optimal window size depends on the density of observations which can be varying for different datasets, and we found the 4-hour window is a suitable value for our dataset. We fill the entry with zero if there is no observation during that 4-hour window. Every 4-hour window is taken within a clinic encounter and can not cross two different encounters to ensure the feature values from different encounters will not be averaged together. For instance, a patient having 2 clinic encounters from 2014-05-05 1:15:00PM to 2014-05-05 6:00:00PM and 2015-09-20 9:12:00AM to 2015-09-20 1:00:00PM will have 4 rows in each table, representing the observations from 2014-05-05 12PM to 2014-05-05 4PM, 2014-05-05 4PM to 2014-05-05 8PM, 2015-09-20 8AM to 2015-09-20 12PM, and 2015-09-20 12PM to 2015-09-20 4PM. Furthermore, we delete a row if it contains all zeros (no observations for any feature). Table \ref{table:EHR} shows the variables we used in our dataset. All patients' demographic data is constructed as a fixed-size vector.

\begin{figure}[ht]
\centering
\includegraphics[width=0.85\textwidth]{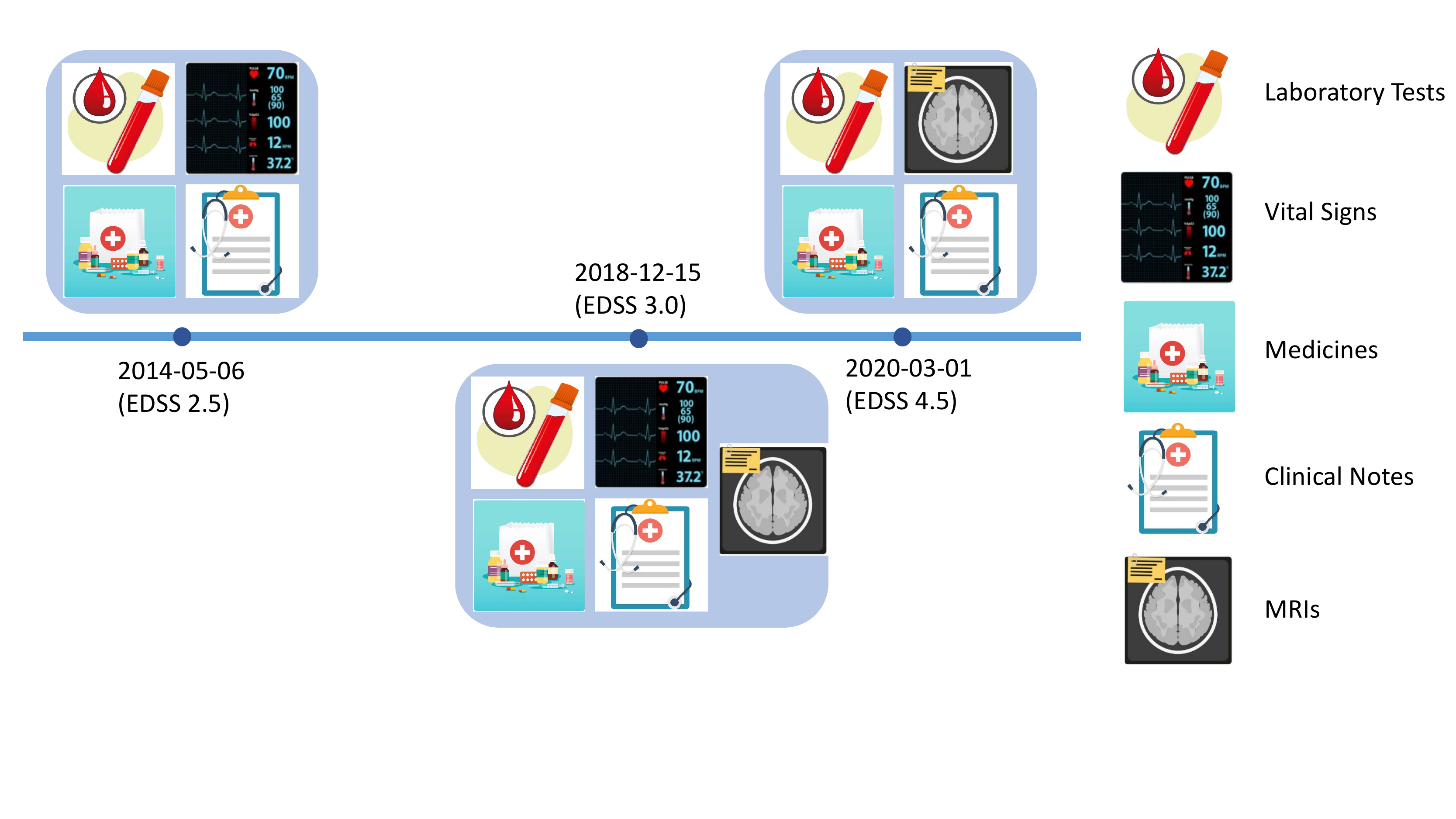}
\caption{The clinic encounters of an example patient and the information (data modality) that were recorded during each clinic visit.}
\label{fig:patientencounters}
\end{figure}

Figure \ref{fig:patientencounters} demonstrates an example patient's three clinic encounters. Note that not all data modalities were observed in each encounter.

\begin{table}[tb]
\caption{The features from the structured EHR data tables, including laboratory tests, vital signs, and medications.}
\resizebox{\textwidth}{!}{
\label{table:EHR}
\begin{tabular}{cccccc}
\hline
\multicolumn{3}{c}{LABORATORY TEST} & VITAL SIGN & MEDICATION  \\ \hline
Mean Corpuscular Hemoglobin & Carbon Dioxide & Albumin &   Diastolic Blood Pressure & Baclofen  \\ 
Red Cell Distribution Width & Basophils & Glucose Level &  Systolic Blood Pressure & Gabapentin  \\ 
Mean Corpuscular Hemoglobin Concentration & White Blood Cell Count & eGFR &  Heart Rate & Copaxone  \\ 
Mean Corpuscular Volume & Hematocrit & Albumin/Globulin Ratio &  Weight & Gilenya  \\ 
Alanine Aminotransferase & Red Blood Cell Count & Eosinophils &   Height & Tecfidera  \\
Aspartate Aminotransferase & Platelet Count & Potassium Level &    BMI & Aubagio  \\ 
Anion Gap & Total Protein & Creatinine &  O2 Saturation & Ampyra  \\ 
MRI Brain W/Wo Contrast & Bili Total & Bilirubin, Direct &  Pulse  & Prednisone \\ 
Creatinine Level & Alkaline Phosphatase & Bun/Creatinine Ratio &   Temperature &  Vitamin  \\ 
Bun/Creatinine Ratio & Albumin Level & Potassium &  Respiration  & Duloxetine  \\ 
Hematocrit Test & Globulin & Systolic &   & Dalfampridine  \\ 
Hemoglobin & Neutrophils & MRI Spine Cervical W Wo Contrast  &    &  Clonazepam \\ 
Blood Urea Nitrogen & Lymphocytes &  Brain W/Wo Contrast MRI &    &   \\ 
Mean Platelet Volume & Absolute Eosinophils & Body Surface Area &    &   \\ 
Calcium Level Total  & Basophils & Bilirubin, Indirect &    &   \\ 
Sodium Level & Absolute Monocytes & Segmented Neutrophils  &    &   \\ 
Thyroid Stimulating Hormone & Absolute Neutrophils & Monocytes &   &   \\ 
Segs-Bands & Absolute Basophils & Chloride Level &    &   \\  \hline
\end{tabular}
}
\end{table}

\section*{An encoder-decoder architecture for data fusion}
We propose a multimodal neural network that takes data in various modalities (structured EHR, clinical notes, MRIs) as input and is trained to predict a patient's EDSS score. The proposed neural network adopts an encoder-decoder schema in a sequential structure with the self-attention module.

\subsection*{Encoder Network}
\begin{figure}[tb]
    \centering
    \includegraphics[width=0.95\textwidth]{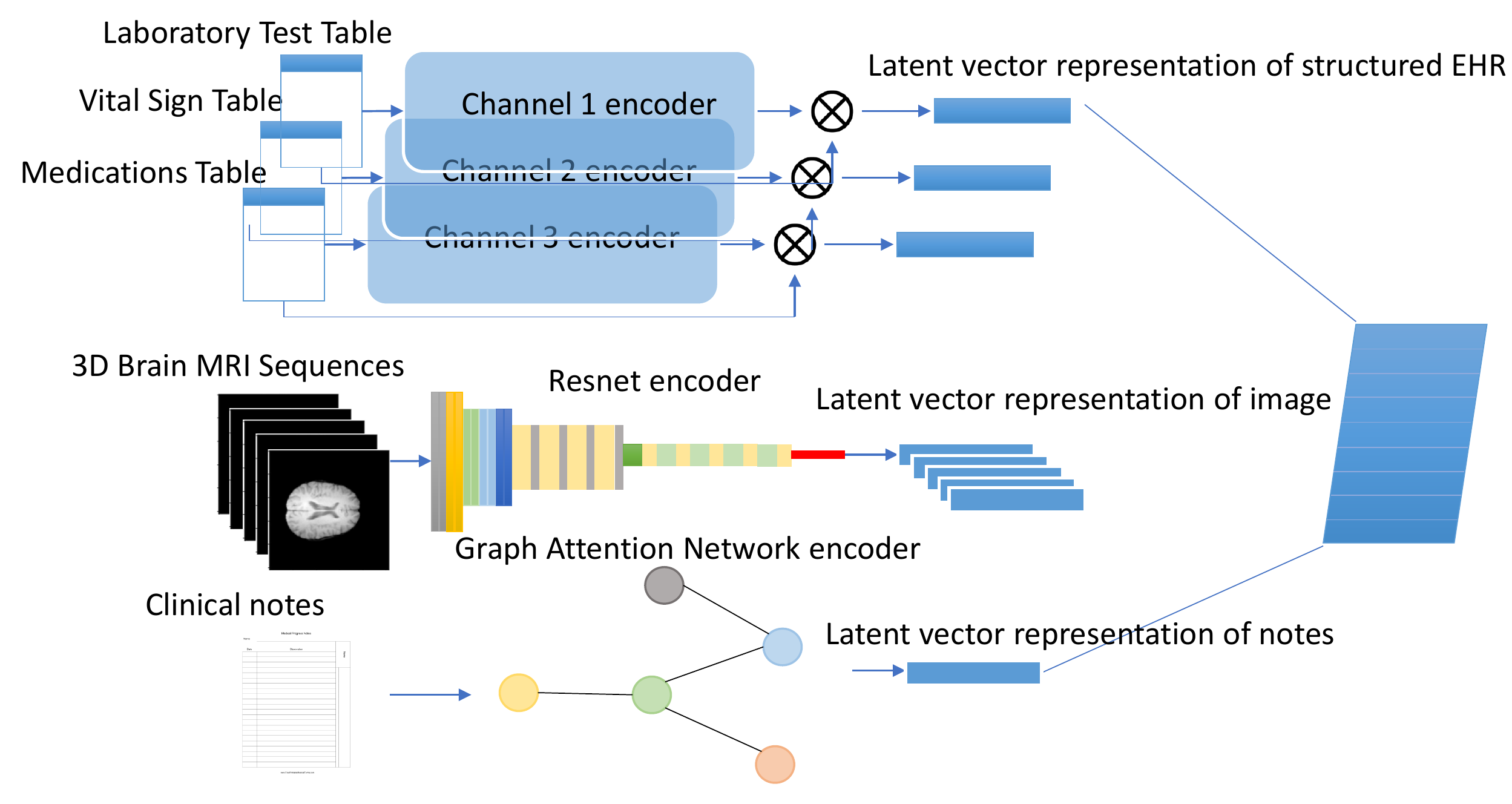}
    \caption{The encoder network for our proposed deep neural network.}
    \label{fig:encoder}
\end{figure}

The goal of the encoder network is to process data of different modalities and transform them into dense embeddings belonging to the same high-dimensional latent space. For different modalities, the neural network employs different encoder neural network architectures that are suitable for the learning task (CNN for image processing and structured EHR, Graph Neural Network or BERT for clinical notes), see Figure \ref{fig:encoder}. 

\begin{figure*}[tb]
    \centering
    \includegraphics[width=0.95\textwidth]{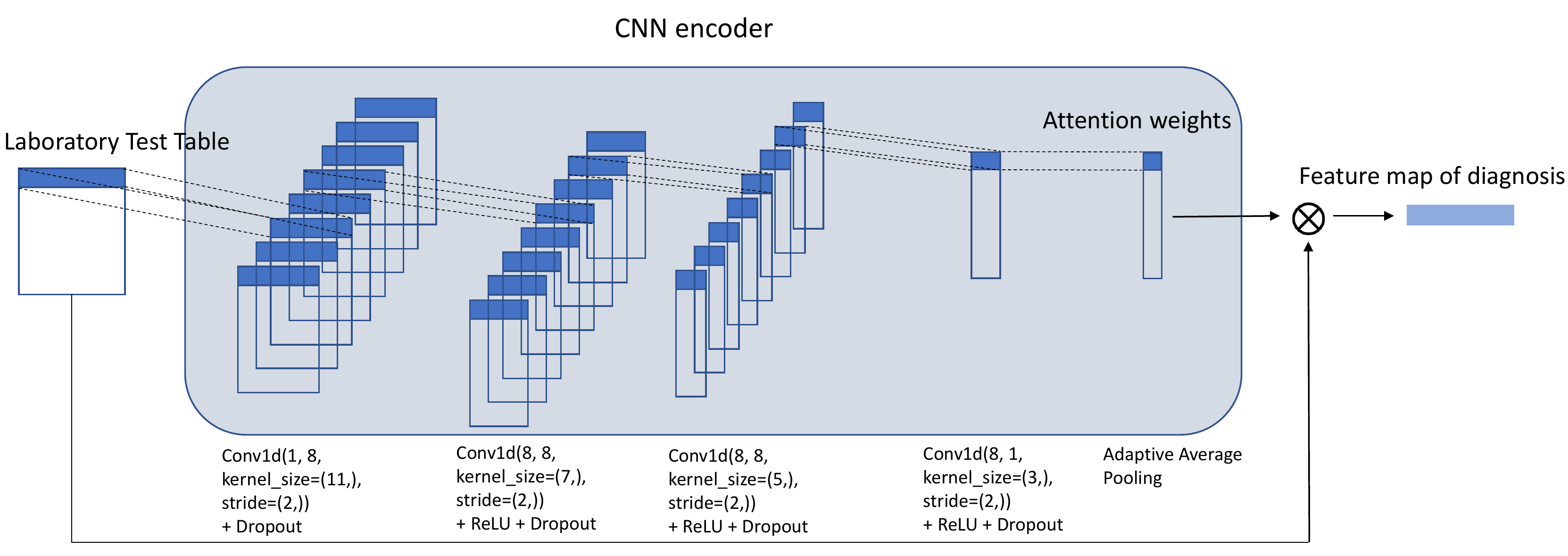}
    \caption{The detailed architecture of one of the encoder channels for processing structured EHR data. The figure shows the lab test channel as an example.}
    \label{fig:encoder-channel}
\end{figure*}

\textbf{Structured EHR.} The encoder network for structured EHR is composed of multiple parallel channels, and different channels embrace homogeneous network structure but with various hyper-parameters to fit the patient's structured EHR table of different sizes, see Figure \ref{fig:encoder-channel}. 

The number of rows for tables can be different across different patients and across different tables of the same patient (if a certain table has no observations in a certain 4-hour window). Such an irregular-sampling format causes heterogeneity along the time axis. Traditional imputation-based methods usually define a shared regular-spaced time axis for all patients and all tables and tries to impute the missing values on the unobserved time points using various techniques, such as filling with the value of zero, average, or majority, etc., forward or backward filling, or advanced techniques such as multiple imputation \cite{van2011mice}, or the Gaussian process \cite{zhang2021real}. Imputation-based methods can be computationally expensive, time-consuming, and most importantly, increase the time dimensions, especially when the time between two clinic encounters is long. 

We introduce a self-attention module into the encoder network to handle the irregularity sampling issue, by promoting the neural network to automatically pay distinctive attention (by assigning different attention weights) to different time points of patient history, aggregating them and producing a vectored representation (embedding) of each table. An attention weight is computed for each row (4-hour time window) by applying multiple layers of 1-dimensional convolutional neural networks (CNNs) on the feature dimension and outputting an attention weight for each time stamp. The attention vector can be seen as attention weights on different time stamps, and after being applied to the original input data, the network will generate an embedding of a fixed dimension that is consistent for all patients.

To be specific, all channels consist of multiple stacked 1D convolution layers followed by the ReLU activation layer and dropout layers. The number of layers is set up differently for different channels according to the number of features in the input tables. For the $i$-th patient, the $k$-th data table $\boldsymbol{D}_{k}^i$ of dimension ${t_{k}^i \times f_k}$ is fed into the $k$-th channel, where $t_k$ rows represent the time stamps of clinic visits and $f_k$ columns represent variables. Note that different EHR tables (laboratory tests, vital signs, medications, etc.) have different $f_k$ and different patients have different numbers of clinic visits $t_{k}^i$. Each row of the table is processed through a stack of multiple 1D CNNs (see Figure  \ref{fig:encoder-channel}) and is reduced to a single value (attention weight). The entire table will generate an attention weight vector $\boldsymbol{\alpha}_{k}^i$ of size ${t_k^{i} \times 1}$. The attention weights can be viewed as the weight factor of all $f_k$ features at different time points. In the following, we omit the patient index $i$.

We multiply the attention vector $\boldsymbol{\alpha}_{k}$ with the input matrix $\boldsymbol{D}_{k}$ to get the feature map $\boldsymbol{e}_{k}$ for each table, 
\begin{equation}
    \boldsymbol{e}_{k} = \boldsymbol{\alpha}_{k}^T \cdot \boldsymbol{D}_{k}.
\end{equation}
where $\boldsymbol{e}_{k}$ is of size $1 \times f_k$. Specifically, each element in $\boldsymbol{e}_{k}$ is calculated as
\begin{equation}
    \boldsymbol{e}_{k}[j] = \sum_{m=1}^{t_k}\boldsymbol{\alpha}_{k}[m]\boldsymbol{D}_{k}[m, :], \textrm{ for } j= 1,\ldots, f_k,
\end{equation}
and $\boldsymbol{e}_{k}$ is the embedding vector of the $k$-th table for a certain patient.

\textbf{Image Embedding.}
The encoder channel for patient MRI images takes a different network structure from the structured EHR. We leverage the ResNet \cite{he2016deep} to process the MRI images. Each MRI sequence (T1-pre, T1-post, T2, PD, FLAIR) is fed into a respective ResNet model. The output is a fixed-length latent representation of each MRI sequence as an embedding vector of a fixed preset dimension. Alternatively, other ResNet variants \cite{xie2017aggregated} could also serve as embedding learning network in our task. Our experiment shows the adoption of different network structure for the MRI sequences brings trivial accuracy improvements on the final prediction performance, due to the reason that 1) the ResNet model itself is powerful enough to capture the key features in the MRI images and generate diverse embeddings for positive and negative patients; 2) the MRI data only accounts for a partial of all the input multimodal data, therefore, the effect of the ResNet variations on the final outcome will be diluted by other data modalities.

\textbf{Clinical Notes Embedding.}
The encoder channel for patient clinical notes data is processed using a graph attention convolution model, which takes text as input and outputs an embedding for each document \cite{nikolentzos2020message}. The medical word embeddings are from a pre-trained database which was trained on PubMed+MIMIC-III \cite{zhang2019biowordvec}. The graph attention model treats the entire document as a word co-occurrence network by representing words in the corpus of all patients' documents as graph nodes. In addition, we add another ``document node'' which represents the entire document and connects to all other nodes. The model maintains a sliding window to capture word co-occurrences, which will be represented as edges of the graph. The edge is directed and weighted, in order to represent the correct word orders in the sliding window and retain maintain meaningful semantics and word co-occurrence counts. The entire network is trained through message passing. We define $G(V,E)$ as the graphical network, and denote a node $v (\in V)$'s neighbors as $\mathcal{N}(v)$. A node $v$ constructs a broadcasting message by aggregating (using multi-layer perceptron) its neighbor node embeddings,
\begin{align}
    \boldsymbol{m}_{v}^{t+1} & =\text { AGGREGATE         }^{t+1}\left(\left\{\boldsymbol{h}_{w}^{t} \mid w \in \mathcal{N}(v)\right\}\right),
\end{align}
which can proceed in a parallel manner using matrix format,
\begin{align}
    \boldsymbol{M}^{t+1} & =\operatorname{MLP}^{t+1}\left(\boldsymbol{D}^{-1} \boldsymbol{A} \boldsymbol{H}^{t}\right),
\end{align}
where $\boldsymbol{H}^{t} \in \boldsymbol{R}^{n \times d}$ is the $d$-dimensional node features of $n$ nodes and $\boldsymbol{A} \in \boldsymbol{R}^{n \times n}$ is the adjacency matrix, and MLP is multiple layer perceptrons neural network.

All nodes update themselves by their own embedding and all messages from their neighbors using a Gated Recurrent Unit (GRU) network,
\begin{align}
    \boldsymbol{h}_{v}^{t+1}=\text { COMBINE }^{t+1}\left(\boldsymbol{h}_{v}^{t}, \boldsymbol{m}_{v}^{t+1}\right),
\end{align}
again in matrix format,
\begin{align}
    \boldsymbol{H}^{t+1}=\operatorname{GRU}\left(\boldsymbol{H}^{t}, \boldsymbol{M}^{t+1}\right).
\end{align}

After $T$ steps, a final self-attention read-out layer is used to aggregate all nodes embeddings and output a latent vector to represent the entire document,
\begin{align}
    \boldsymbol{Y}^{T} &=\tanh \left(\hat{\boldsymbol{H}}^{T} \boldsymbol{W}_{A}^{T}\right) \\
    \boldsymbol{\beta}_{i}^{T} &=\frac{\exp \left(\boldsymbol{Y}_{\boldsymbol{i}}^{T} \cdot \boldsymbol{v}^{T}\right)}{\sum_{j=1}^{n-1} \exp \left(\boldsymbol{Y}_{\boldsymbol{j}}^{T} \cdot \boldsymbol{v}^{T}\right)} \\
    \boldsymbol{u}^{T} &=\sum_{i=1}^{n-1} \boldsymbol{\beta}_{i}^{T} \hat{\boldsymbol{H}}_{i}^{T}
\end{align}
where $\hat{\boldsymbol{H}}^{T} \in \boldsymbol{R}^{n \times d}$ is the final node representation of all $n-1$ nodes (remove the document node) after $T$ time steps, and $\boldsymbol{W}_{A}^{T}$ is the network parameters (a dense layer). Therefore, $\boldsymbol{u}^{T} \in \boldsymbol{R}^{d}$ would be the final representation of the document, i.e. aggregation of all node features, which will be fed into a classification layer for document classification.

\subsection*{Multi-modality Fusion}

Medical data often have multiple types of information (demographics, vital, lab, diagnosis, procedure, medication, etc.) and there is intrinsic logic behind them. For example, vitals, and labs contribute to the diagnosis, and diagnosis will determine the procedure and medication. Some of this information is temporally invariant (e.g., demographics) and others are changing over time. Therefore, they need to be handled differently. Based on the causal relationships (vital, lab, MRI scan) $\rightarrow$ (diagnosis) $\rightarrow$ (prescription, procedure) $\rightarrow$  (medicine administration), we build our data fusion pipeline for time-variant information through the aforementioned bidirectional GRU-based decoder. The order of the inputs (see the left part of Figure  \ref{fig:decoder}) are organized in a way to learn the intrinsic relationships of such information.  The latent representation vectors from each encoder network channel are stacked into a regular matrix $\boldsymbol{E}$ (zero-padded if not the same length), where each row represents a modality
\begin{equation}
    \boldsymbol{E} = [\text{ZeroPadding}(\boldsymbol{e}_1)^T, \ldots, \text{ZeroPadding}(\boldsymbol{e}_K)^T]^T,
\end{equation}
where $\boldsymbol{E}$ is of dimension $K \times d, d=\max(f_1, \ldots, f_K)$.

We integrate the time-invariant demographics at the end of the layer as a late fusion step (see the right part of Figure  \ref{fig:decoder}) to combine both pieces of information holistically. 

\subsection*{Decoder Network}
We propose a decoder network structure that is composed of a stacked bidirectional GRU (Bi-GRU) network with a self-attention module taking the feature matrix $\boldsymbol{E}$ as input. The self-attention serves to learn important weights on the state vectors from different data modalities. The Bi-GRU network takes $K$ as the sequence length and $d$ as the input size. We use $\boldsymbol{C}$ to denote the stack of hidden states of all time points, which is of dimension $K \times h, h=2 \times \textrm{hiddensize}$ (note that factor 2 comes from the bi-direction network being used).

\begin{figure*}[tb]
    \centering
    \includegraphics[width=0.95\textwidth]{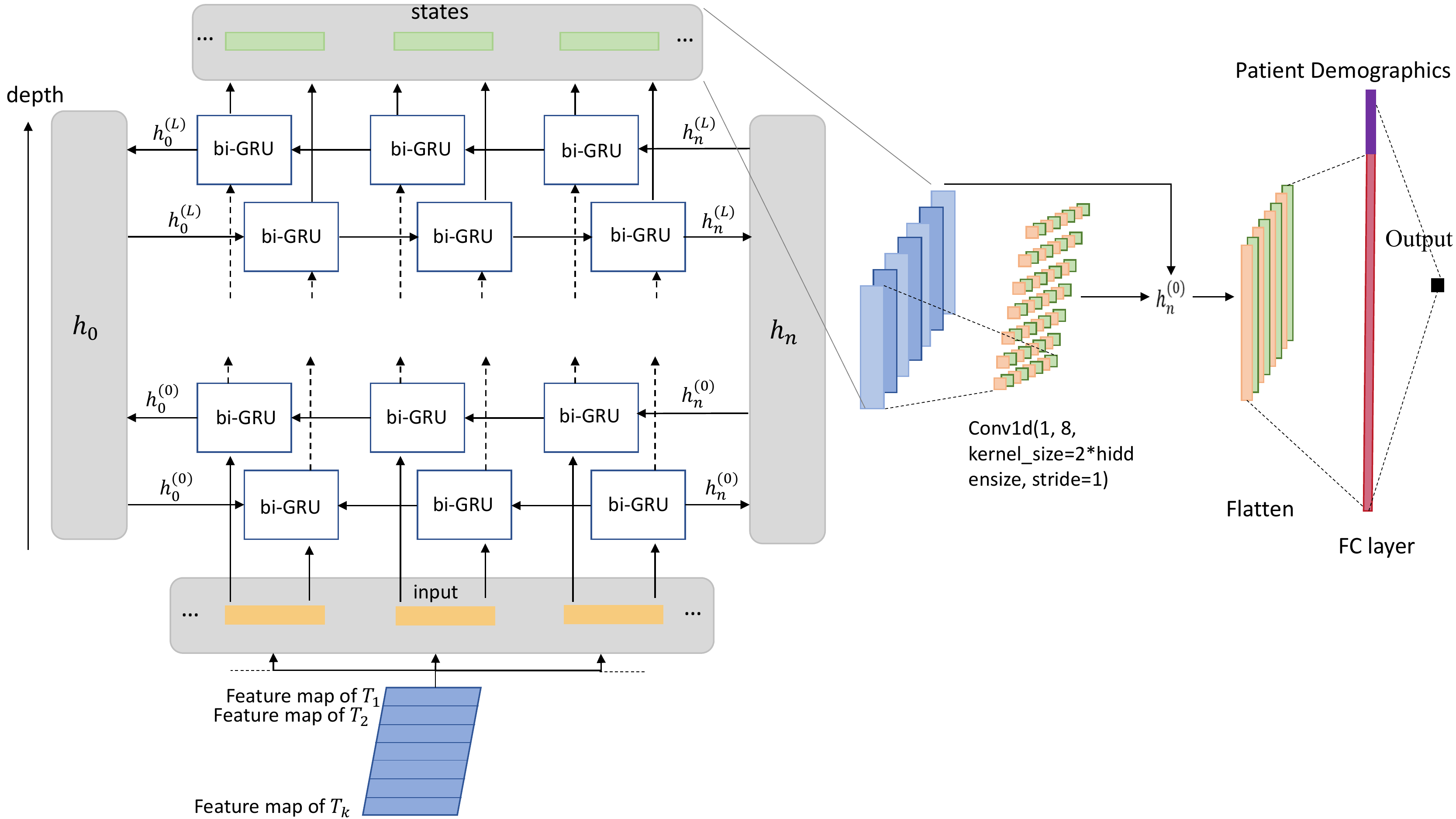}
    \caption{The decoder network for our proposed deep neural network.}
    \label{fig:decoder}
\end{figure*}

Each state of the bidirectional GRU network is fed into an attention module, which is 1D convolution layer of multiple output channels. The attention module outputs a vector of attention weights $\boldsymbol{\gamma}$ of length $g$ (hyper-parameter, depending on the output channel of the convolution layer), and
\begin{equation}
    \centering
    \boldsymbol{B} = [\boldsymbol{\gamma}_1^T, \ldots, \boldsymbol{\gamma}_K^T]^T,
\end{equation}
where $\boldsymbol{B}$ is of dimension $K \times g$ denoting the attention matrix. The attention matrix is multiplied with the GRU output,
\begin{equation}
    \boldsymbol{O} = \boldsymbol{B}^T \cdot \boldsymbol{C}.
\end{equation}
where $\boldsymbol{O}$ is of dimension $g \times h$. Note that the purpose of this attention layer is to enforce a feature reduction from the high-dimensional GRU outputs to a smaller and more informative lower-dimensional embedding not only for reducing the noise but also for increasing the efficiency of neural network training.

The output matrix $ \boldsymbol{O}$ is flattened, and concatenated with the patient demographic data vector $\boldsymbol{d}$, and fed into a fully-connected (FC) layer for prediction,
\begin{equation}
    o = \text{FC}(\text{Concat}(\text{Flatten}(\boldsymbol{O}), \boldsymbol{d})).
\end{equation}
see Figure  \ref{fig:decoder}.

\section*{Experiments}
\label{sec-experiments}
\subsection*{MRI Images}
We introduce five channels to process the MRI sequence, where each channel employs a ResNet structure. The five channels are independent and each is trained to learn from one sequence (T1-pre, T1-post, T2, FLAIR and PD). All MRI images are bias-corrected, skull-stripped, and registered and the intensity scale is normalized \cite{nyul1999standardizing}. If a patient performed MRI scans in more than one hospital visit, we use the last scan as it represents the patient's most recent disease status. Due to the relatively high imbalance of the positive and negative samples, we performed 10-fold re-sampling for the negative training samples during model training. To add robustness to the learning of ResNet, all input MRI images are also randomly rotated with a probability of 0.5 by a maximum of $\pm 0.02$ degrees on all three dimensions.  For each channel, a respective ResNet model is trained on the training dataset, and we select the trained model with the best performance on the validation dataset. Our goal is to learn a latent vector representation instead of performing disease classification, therefore, the training process is formulated as a metric learning task where each channel's ResNet is trained to learn an embedding for each MRI sequence of a patient. The triplet margin loss \cite{balntas2016learning} operates directly on embedding distances by promoting the matching point (positive) to the reference point (anchor) and the non-matching point (negative) away from the anchor. By using a triplet margin loss, the network learns well-separated embedding vectors for positive and negative patients for downstream decoding networks to perform classification. The triplet margin loss is defined as

\begin{figure*}[tb]
    \centering
    \includegraphics[width=0.95\textwidth]{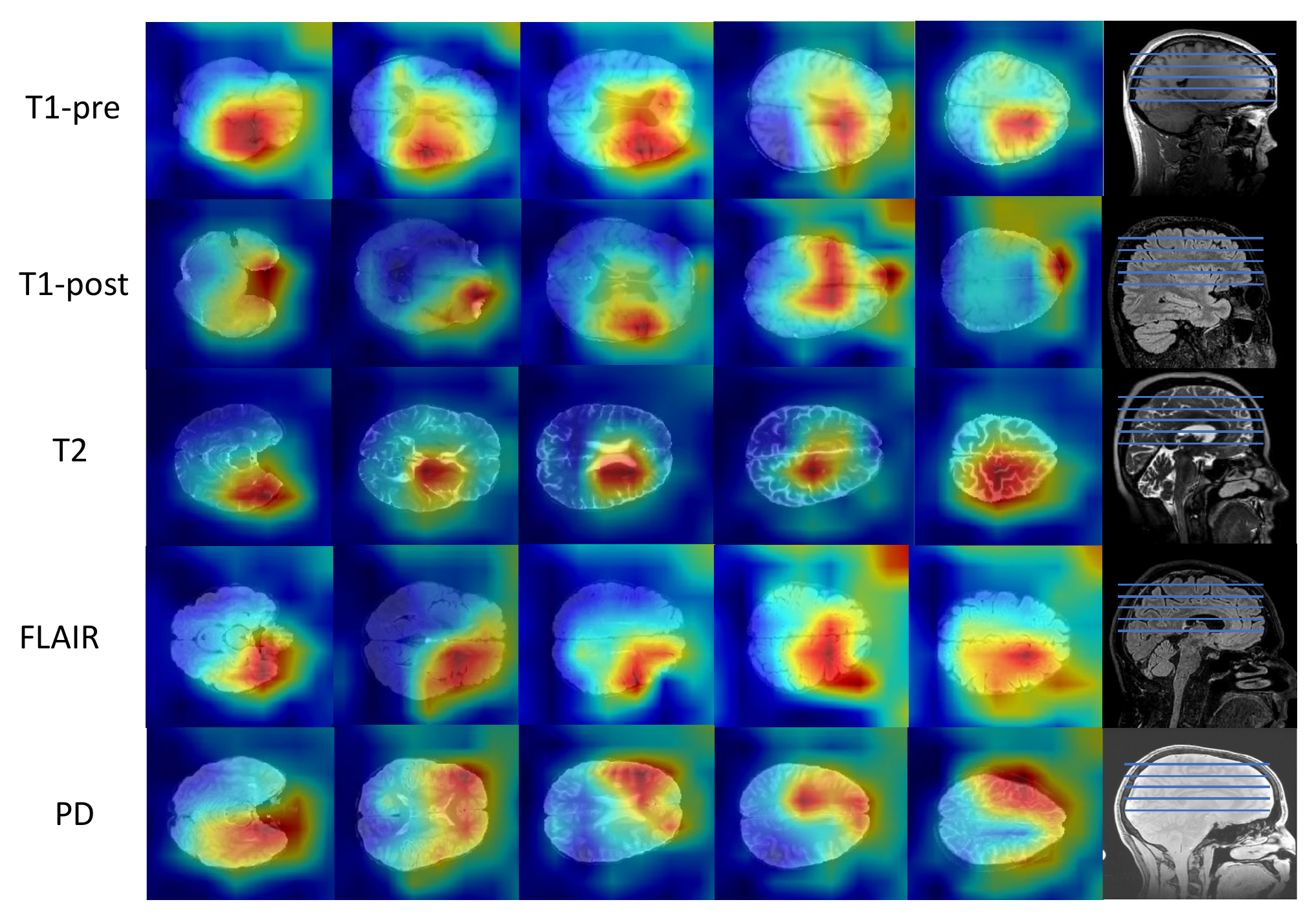}
    \caption{Attention maps for MRI sequences of a sample patient.}
    \label{fig:MRI_attention}
\end{figure*}

\begin{align}
\label{eq:marginloss}
    loss= \sum_{a_i, n_i, p_i \in \text{batch}} \max \left( d\left(a_{i}, p_{i}\right)-d\left(a_{i}, n_{i}\right)+\operatorname{margin}, 0\right),
\end{align}
where $a_i, p_i$, and $n_i$ are an anchor, positive and negative sample in the batch, respectively. We set the anchor point in our model as a fixed point in the embedding space, therefore, the distance from the positive samples to the anchor is minimized and the distance from the negative samples to the anchor is maximized. 

The margin in the triplet margin loss is chosen to be 1.5. The learning rate is set to be $10^{-5}$ and the batch size is 10. The ResNet in each encoder channel is trained for 500 epochs. Early stopping criteria of not-improving for consecutive 50 epochs on the validation dataset is adopted.

We leverage the gradient-weighted class activation mapping (Grad-CAM) \cite{selvaraju2017grad} model to locate and visualize the important regions the ResNet neural network is learning for predicting the target. The Grad-CAM uses flowing gradients of the prediction target into the last convolutional layer of the ResNet to produce a heatmap of the regions according to their contributions to the prediction, see Figure \ref{fig:MRI_attention}, 

\subsection*{Clinical Notes}
We preprocess patients' clinical notes by identifying and then removing all sensitive patient health information that is irrelevant to our prediction task, including the patient and physician's name, address, phone number, and email address. Similar to the MRI image data, we formulate the embedding generation problem from clinical notes as a metric learning problem, where the message-passing graph neural network is trained to learn meaningful embeddings and their distances between positive and negative samples. Hence, the same loss function (\ref{eq:marginloss}) is used for this encoder channel.

We set the size of the window to be 10 (covering 10 consecutive words) and the message passing layer to be 2. The hidden side of the GRU network is 64. We trained the graph network with 500 epochs with a batch size of 128, the learning rate of $10^{-3}$, and early stopping criteria of 50 epochs (no improvements on the validation dataset). We choose the best-performing model on the validation dataset and run it on the test dataset to get the model's final performance.

\subsection*{Structured EHR}
The patient's structured EHR consists of tables of 4 categories, the laboratory tests table, the vital signs table, the medications table, and the demographics table.  The first 3 categories are in the format \textit{number of timestamps} $\times$ \textit{number of features} containing the laboratory test results (float), vital sign measurements (float), and medications (0/1 indicators), respectively. Table \ref{table:EHR} shows a pre-selected subset of all the variables from the above 3 categories to be used in our model, based on their observation frequency. The demographic table contains race (0/1, one-hot encoded), ethnicity (0/1, one-hot encoded), sex (0/1, male/female), and age (float, min-max normalized). The encoder network consistents 3 channels for each of the first 3 categories, and the network parameters are described in Table \ref{table:encoder-parameters}. 

\begin{table*}[tb]
\caption{Encoder network parameters (I: input channel size, O: output channel size, K: kernel size, S: stride size, P: padding size, R: (dropout) rate). }
\label{table:encoder-parameters}
\begin{adjustbox}{width=\textwidth ,center}
\begin{tabular}{llllllllllll}
\toprule
& Conv1d    & Dropout & Conv1d  & ReLU + Dropout & Conv1d   & ReLU + Dropout  & Pooling \\ 
\midrule
Channel 1 (Lab tests)         & I: 1, O: 8, K: 7, S: 2 & R: 0.3  & I: 8, O: 8, K: 4, S: 2 & R: 0.3         & I: 8, O: 1, K: 3, S: 2 & R: 0.3    & Avg.    \\ 
Channel 2 (Vital Sign Observ.) & I: 1, O: 8, K: 3, S: 2  & R: 0.3  & I: 8, O: 1, K: 2, S: 2  & R: 0.3       &  - & -   & Avg.    \\ 
Channel 3 (Medication)  & I: 1, O: 8, K: 3, S: 2 & R: 0.3  & I: 8, O: 1, K: 2, S: 2 & R: 0.3         & -  & -    & Avg.    \\ 
\bottomrule
\end{tabular}
\end{adjustbox}
\end{table*}

A patient's three structured EHR's embeddings produced by the encoder network will be concatenated with the five MRI image embeddings produced by the ResNet, and together with the clinical note embedding to be fed into the decoder network. In the situation of a patient (a small amount) without MRI or clinical notes, the corresponding embedding will be set to an all-zero vector.  In the decoder network, the bidirectional GRU network is set to have 4 layers and hidden size of 512.

\subsection*{Results}
We used 5-fold cross-validation by randomly split the 300 patients into five folds and iteratively using each fold as the hold-out test set (20\%) and the remaining as the training set (80\%). The model's performance of predicting EDSS $> 4.0$ using different data modalities and their combinations are presented in Table \ref{table:result}. The prediction goal is whether the patient's EDSS $> 4$ of the current clinic visit, using longitudinal data of the current clinic visit and all previous visits. Multimodal data inputs in general perform much better than single modal input, and the top-3 AUROC performances are when all data (0.9301), EHR \& Notes (0.9196), and MRIs \& Notes (0.9201) are being used. The degradation in performance by deleting MRI or EHR information from the input data is very limited. However, if clinical notes were to be deleted, the performance drops to 0.8499. Table \ref{table:result_milestone} shows the model's performance for predicting other EDSS milestones (EDSS$>$6 and EDSS$>$7) using all data modalities.

\begin{figure*}[tb]
    \centering
    \includegraphics[width=1.0\textwidth]{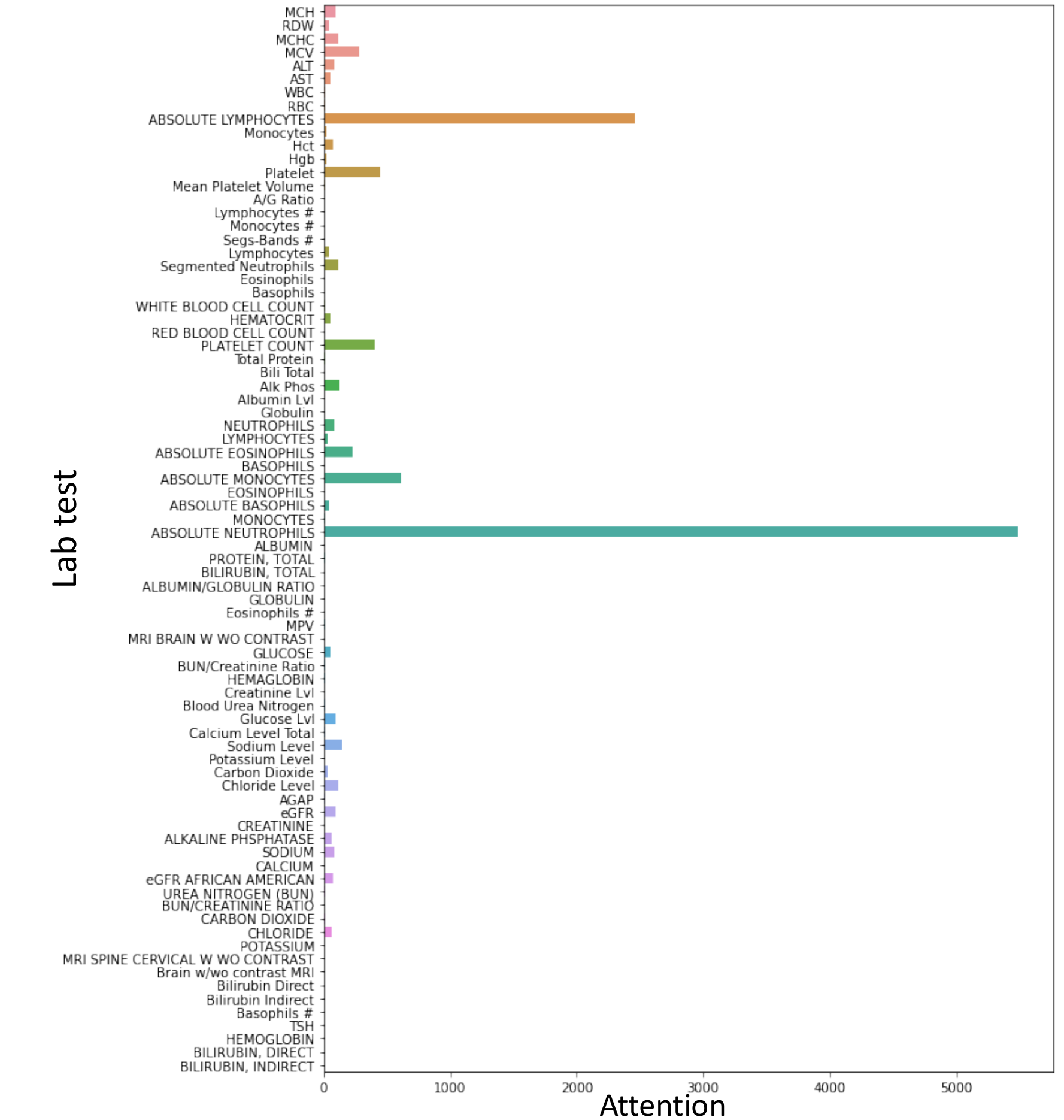}
    \caption{The attention weights for laboratory tests.}
    \label{fig:attention_1}
\end{figure*}

\begin{figure*}[tb]
    \centering
    \includegraphics[width=0.85\textwidth]{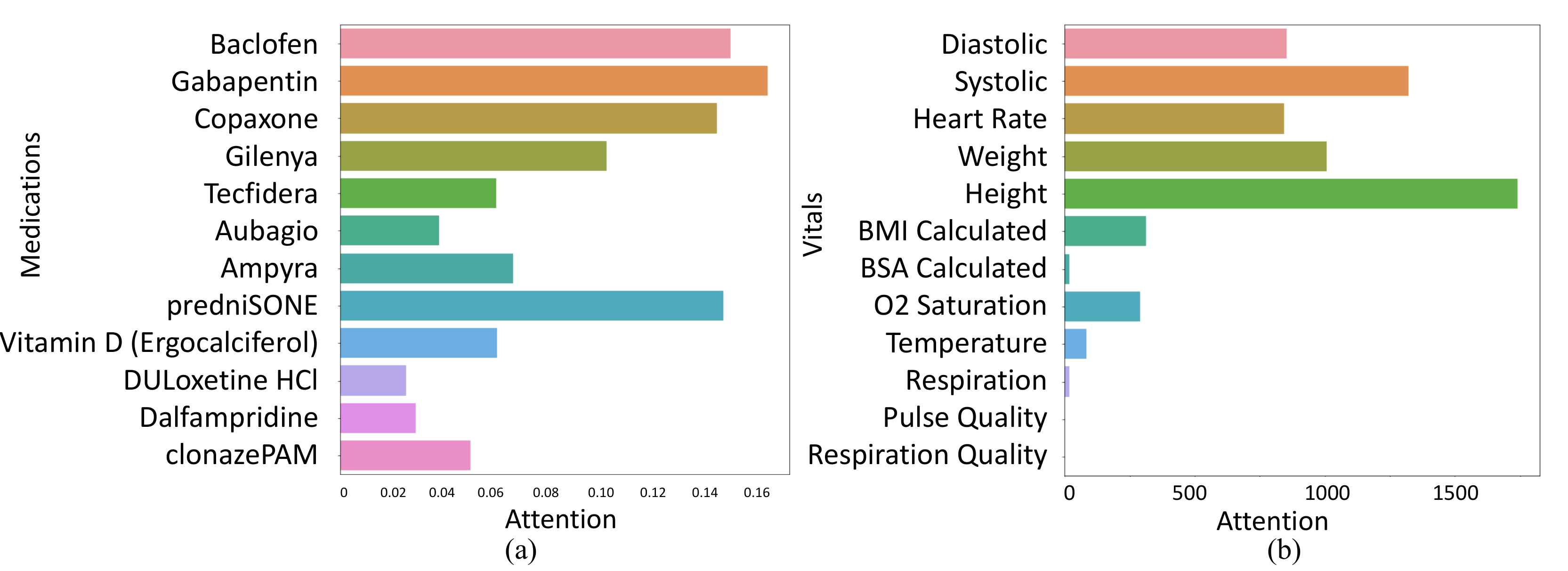}
    \caption{The attention weights for (a) medications and (b) vital signs.}
    \label{fig:attention_2}
\end{figure*}

The encoder channels (laboratory tests, vital signs, medications) for the structured EHR, as a self-attention network by itself, could also generate the feature importance. Notice that the feature importance can be generated both on an individual-level and global level. The latter is evaluated as the average of all feature importance of all individuals. Figure \ref{fig:attention_1} shows the global importance of all laboratory features evaluated on all patients in the test set, where a larger value corresponds to the higher importance of a feature. The top 3 important features for all patients, as can be seen from the figure, are ``Absolute Neutrophils'', ``Absolute Lymphocytes'' and ``Absolute Monocytes''. Figure \ref{fig:attention_2} shows the global-feature importance on all vital signs and medications, some medicine such as the ``Baclofen 10 MG Oral Tablet'', ``Gabapentin 300 MG Oral Capsule'', ``predniSONE 50 MG Oral Tablet'' were commonly used to treat MS symptoms which were identified and assigned with high importance by our algorithm. For the attention weights on vital signs, the ``Temperature'', ``Respiration'', ``Pulse Quality'' and ``Respiration Quality'' are reasonably assigned with the least importance for the prediction of MS. 

\begin{table*}[tb]
\caption{Prediction accuracy performance of using different data modalities for predicting EDSS$>$4. In each evaluation metric, the top-3 highest scores are highlighted. }
\label{table:result}
\resizebox{\columnwidth}{!}{%
\begin{tabular}{lcccccc}
\hline
               & AUROC  & AUPRC  & Sensitivity & Specificity & F1 score & Accuracy \\ \hline
MRI T1-pre     & $0.6749 \displaystyle \pm 0.0326$  & $0.2372 \displaystyle \pm 0.0325$   & $0.5343 \displaystyle \pm 0.0436$  & $0.7123 \displaystyle \pm 0.0326$   & $0.2305 \displaystyle \pm 0.0225$  & $0.6908 \displaystyle \pm 0.0398$  \\
MRI T1-post    & $0.6824 \displaystyle \pm 0.0488$ & $0.2213\displaystyle \pm 0.0298$ & $0.6012 \displaystyle \pm 0.0456$     & $0.7041 \displaystyle \pm 0.0325$     & $0.2562 \displaystyle \pm 0.0399$  & $0.6912  \displaystyle \pm 0.0222$ \\
MRI T2         & $0.8195 \displaystyle \pm 0.0333$ & $0.2554 \displaystyle \pm 0.0210$ & $0.7323  \displaystyle \pm 0.0255$   & $ 0.7794  \displaystyle \pm 0.0432$    & $ 0.3601 \displaystyle \pm 0.0228$  & $ 0.7742 \displaystyle \pm0.0265$  \\
MRI FLAIR      & $0.8346 \displaystyle \pm 0.0369$ & $0.3500 \displaystyle \pm 0.0399$ & $\mathbf{0.8001 \displaystyle \pm 0.0280}$     & $0.7776 \displaystyle \pm 0.0298$     & $0.3868 \displaystyle \pm 0.0301$  & $0.7823 \displaystyle \pm 0.0487$  \\
MRI PD         & $0.5870 \displaystyle \pm 0.0445$ & $0.1024 \displaystyle \pm 0.0380$ & $\mathbf{0.8016  \displaystyle \pm 0.0209}$    & $0.5262  \displaystyle \pm 0.0251$    & $0.2346 \displaystyle \pm 0.0244 $ & $0.5495 \displaystyle \pm 0.0477$  \\
Clinical Notes & $0.7385 \displaystyle \pm 0.0422$ & $0.5560 \displaystyle \pm 0.0303$ & $0.5023 \displaystyle \pm 0.0280  $    & $\mathbf{0.9461 \displaystyle \pm 0.0362}$     & $\mathbf{0.8813 \displaystyle \pm 0.0312}$   & $0.5467 \displaystyle \pm 0.0399 $ \\
Structured EHR & $0.6981 \displaystyle \pm 0.0254 $ & $0.3821 \displaystyle \pm0.0086$ & $0.7517 \displaystyle \pm 0.0347$    & $0.6979 \displaystyle \pm 0.0433 $    & $0.5064 \displaystyle \pm 0.0276 $ & $0.7097  \displaystyle \pm 0.0389$ \\
MRIs \& Notes   & $\mathbf{0.9201 \displaystyle \pm 0.0268}$ & $\mathbf{0.8748 \displaystyle \pm 0.0365}$ & $0.7513  \displaystyle \pm0.0347$    & $\mathbf{0.9780 \displaystyle \pm 0.0312} $  & $\mathbf{0.8195 \displaystyle \pm 0.0388}$  & $\mathbf{0.9346 \displaystyle \pm 0.0245}$ \\  
MRIs \& EHR     & $0.8499 \displaystyle \pm 0.0400$ & $0.4404\displaystyle \pm 0.0301$  & $0.7128 \displaystyle \pm 0.0382$     & $0.7108 \displaystyle \pm 0.0369$  & $0.4978 \displaystyle \pm 0.0487$ & $0.7121\displaystyle \pm 0.0210$  \\
EHR \& Notes    & $\mathbf{0.9196 \displaystyle \pm 0.0368}$ & $\mathbf{0.8448 \displaystyle \pm 0.0243}$ & $0.7527 \displaystyle \pm 0.0265 $   & $0.9165 \displaystyle \pm 0.0245$ & $0.7436 \displaystyle \pm 0.0431 $ & $\mathbf{0.8959 \displaystyle \pm  0.0265}$\\
MS-BERT(\cite{costa2020multiple})  & $0.6600 \displaystyle \pm 0.0310$ &$ 0.2681  \displaystyle \pm 0.0214$    & $0.3330   \displaystyle \pm 0.0312$   & $0.8800 \displaystyle \pm 0.0442$ & $0.2860 \displaystyle \pm 0.0441$  & $0.8210 \displaystyle \pm 0.0420$  \\ 
MRI \& Notes \& EHR  & $\mathbf{0.9301 \displaystyle \pm 0.0423}
$ &$\mathbf{ 0.8623  \displaystyle \pm 0.0312}$    & $\mathbf{0.7899   \displaystyle \pm 0.0265}$   & $\mathbf{0.9790 \displaystyle \pm 0.0214}$ & $\mathbf{0.8114 \displaystyle \pm 0.0315}$  & $\mathbf{0.9354 \displaystyle \pm 0.0365}$  \\ \hline
\end{tabular}
}
\end{table*}

\begin{table*}[tb]
\caption{Prediction accuracy performance at different EDSS milestones.}
\label{table:result_milestone}
\resizebox{\columnwidth}{!}{%
\begin{tabular}{lcccccc}
\hline
               & AUROC  & AUPRC  & Sensitivity & Specificity & F1 score & Accuracy \\ \hline
MRI \& Notes \& EHR (EDSS$>$4) & $0.9301 \displaystyle \pm 0.0423
$ &$ 0.8623  \displaystyle \pm 0.0312$    & $0.7899   \displaystyle \pm 0.0265$   & $0.9790 \displaystyle \pm 0.0214$ & $0.8114 \displaystyle \pm 0.0315$  & $0.9354 \displaystyle \pm 0.0365$ \\ 
MRI \& Notes \& EHR (EDSS$>$6) & $0.9101 \displaystyle \pm 0.0358$ &$ 0.7419  \displaystyle \pm 0.0651$    & $0.8460   \displaystyle \pm 0.0485$   & $0.8043 \displaystyle \pm 0.0364$ & $0.9089 \displaystyle \pm 0.0517$  & $0.6893 \displaystyle \pm 0.0601$  \\ 
MRI \& Notes \& EHR (EDSS$>$7) & $0.9771 \displaystyle \pm 0.0210$ &$ 0.9166  \displaystyle \pm 0.0615$    & $0.6671   \displaystyle \pm 0.0356$   & $0.9913 \displaystyle \pm 0.0214$ & $0.9836 \displaystyle \pm 0.0401$  & $0.6674 \displaystyle \pm 0.0365$  \\ \hline
\end{tabular}
}
\end{table*}

\section*{Discussion}
\label{sec-discussion} 
We propose a multimodal deep neural network that harnesses EHR and neuroimaging to tackle the MS disease severity prediction problem. Our approach embraces rich information from data of multimodality, including laboratory tests, vital signs, medications, neuroimaging data, and clinical notes to provide the EDSS score, which is a commonly used metric to evaluate MS disease severity. This study serves as a first step in leveraging multiple data sources for MS severity prediction, and exploring the effectiveness of each modality in terms of MS prediction. Our experiments show the most useful information for MS severity prediction is embedded in the brain MRI images and the clinical notes, and structured EHR contributes the least to this prediction goal. 

The results shows that despite the many publications, conventional MRI contains relatively less information about MS severity compared to other data modalities. However, T2 and Flair MRI performed relatively better than other MRI sequences. Clinical notes were well-documented to be used for the prediction of EDSS, which has been re-verified in our experiment as the relatively not good performance of using MRIs, or EHR, or MRIs \& EHR were all improved when clinical notes were added to the input. A re-examination of the data reveals a reasonable explanation that the clinical notes contain rich patient general disease information including patient status, medical procedures, and treatment information, which implicitly and partially embeds information from the EHR data and MRI images. 

\subsection*{Future research directions and limitations} The study focuses on predicting patients' MS severity as current clinic visits by using current and historical medical information, with the goal to develop an AI-based patient disease status evaluation tool to replace the human expert. A more interesting research question is to predict a patient's MS disease progression in the future. This necessarily needs to consider the EDSS change ``rate'' by considering the disease duration. For example, having an EDSS of 4.0 at age of 65 and a disease duration of 40 years would mean a relatively benign disease but having an EDSS score of 4.0 only after 5 years of MS diagnosis is considered as ``aggressive'' MS. Moreover, as one of the reviewers that has helped improve this paper pointed out, the severity of MS can be seen as a relative concept instead of an absolute one. The severity of MS should be studied based on an understanding of the ``natural'' disease progression, and it varies in terms of many factors (eg. sex, disease duration, lesion load, atrophy, etc.) Limited by the data size and commonly agreed on criteria to distinguish the ``aggressive'' cases from the rest, we focus on developing a tool to predict EDSS milestones at the moment and leave the decision of MS severity to MS specialists by jointly considering all the above factors. In addition, this problem itself is quite an interesting research problem and could potentially be studied using survival methods, the results will have a high impact on the prevention of rapid disease progression through early intervention.

The second is the limitation of the imaging data. While random rotation of MRI scans (a data augmentation technique used to train ResNet on the MRI sequences) helps generalizability, the use of only one scanner for all datasets makes it difficult to infer if the model would work in the same way when introduced to new images from a different scanner. Therefore, our work serves as a proof-of-concept regarding this question. Ideally, more data (especially data from external sources) needs to be collaboratively collected to verify the inclusion of MRI potentially has a positive impact on a multi-modal model.

Thirdly, the study was conducted on a cohort of 300 MS patients from a local academic medical center. An important future research direction is to evaluate the generalizability of the proposed model to other institutions. The result replicability should be checked from two perspectives, the first is the prediction accuracy with or without model retraining, and the second is if the ranking of importance for different data modalities is the same in general, for example, MRI images and clinical notes contains more signals compared to the structured EHR. If the results in this study are verified, it may serve as a cost-effective study recommending which electronic health information should be collected to reach maximum prediction accuracy.

\section*{Abbreviations}
MS: Multiple sclerosis;
MSSS: Multiple sclerosis severity scores;
P-MSSS: Patient-derived MSSS;
EDSS: Expanded disability status scale;
EHR: Electronic health records;
AUROC: Area Under the receiver operating characteristic curve;
AUPRC: Area under the precision-recall curve
MRI: Magnetic resonance imaging;
SD: standard deviation;
PD: Proton density;
GRU: Gated Recurrent unit;
Grad-CAM: Gradient-weighted class activation mapping; 
BCE: binary cross entropy.

\section*{Supplementary Information}



\begin{backmatter}

\section*{Acknowledgements}
We thank the reviewers for proposing a critical perspective of viewing this research problem which greatly helped improve the quality of this manuscript.

\section*{Authors' contributions}
KZ, SS, and XJ conceived the original idea and designed the model. KZ implemented the model and conducted the experiments. EB and JL contributed to the conception, data acquisition, and interpretation. KZ formed the manuscript. SS, EB, JL, and XJ critically revised the manuscript. EB and JL were in charge of overall direction and planning and helped supervise the project.

\section*{Funding}
XJ is CPRIT Scholar in Cancer Research (RR180012), and he was supported in part by Christopher Sarofim Family Professorship, UT Stars award, UTHealth startup, the National Institute of Health (NIH) under award number R01AG066749 and U01TR002062,

\section*{Availability of data and materials}
The data that support the findings of this study are available on request from the corresponding author SS. The data are not publicly available due to their containing information that could compromise the privacy of research participants. Code is publicly available on Github: https://github.com/anotherkaizhang/MS.

\section*{Declarations}
\section*{Ethical approval and consent to participate}
The study protocol was approved before the initiation of this study by the Committee for the Protection of Human Subjects of the University of Texas Health Science Center at Houston under IRB: HSC-MS-02-090. All recruited patients provided written informed consent upon enrollment. All methods were performed in accordance with the Declarations of Helsinki.

\section*{Consent for publication}
Not applicable.

\section*{Competing interests}
The authors declare that there is no conflict of interest.



\bibliographystyle{bmc-mathphys} 
\bibliography{bmc_article}      


\end{backmatter}
\end{document}